\pdfoutput=1
\documentclass{article}
\usepackage{spconf,amsmath,graphicx}
\usepackage{todonotes}
\usepackage{dsfont}
\usepackage{subcaption}
\usepackage{hhline}

\DeclareMathOperator*{\argmin}{arg\,min}

\title{Coupled Representation Learning for Domains, Intents and Slots in Spoken Language Understanding}

\name{Jihwan Lee, Dongchan Kim, Ruhi Sarikaya, Young-Bum Kim}
\address{Amazon Alexa\\
	Seattle, Washington, USA\\
	\texttt{\{jihwl, dongchan, rsarikay, youngbum\}@amazon.com}}
\begin{document}
%
\maketitle
\begin{abstract}

Representation learning is an essential problem in a wide range of applications and it is important for performing downstream tasks successfully. In this paper, we propose a new model that learns coupled representations of domains, intents, and slots by taking advantage of their hierarchical dependency in a Spoken Language Understanding system. Our proposed model learns the vector representation of intents based on the slots tied to these intents by aggregating the representations of the slots. Similarly, the vector representation of a domain is learned by aggregating the representations of the intents tied to a specific domain. To the best of our knowledge, it is the first approach to jointly learning the representations of domains, intents, and slots using their hierarchical relationships. The experimental results demonstrate the effectiveness of the representations learned by our model, as evidenced by improved performance on the contextual cross-domain reranking task.
\end{abstract}
%


\section{INTRODUCTION}
\label{sec:introduction}

Recent changes from touch to speech as the primary user interface in diverse computing devices has led to tremendous growth of voice-driven intelligent personal digital assistants, such as Amazon Alexa, Google Assistant, Apple Siri and Microsoft Cortana~\cite{sarikaya-IEEE-SPM-paper}. Spoken Language Understanding (SLU) is a key component of these systems, which interprets the transcribed human speech directed at these systems. The main tasks of SLU are domain classification, intent determination, and slot filling\cite{kim2017onenet}. Many researchers in the literature have proposed different approaches that leverage information on one of these tasks to improve the other task~\cite{kim2017onenet, kim2016intent, wen2017jointly, xu2013convolutional, mesnil2015using, Liu2016AttentionBasedRN,kim2017speaker,kim2016frustratingly,kim2017adversarial,kim2017domain,kim2015new,kim2015weakly} and they have shown remarkable performance gains. However, the problem of representation learning for domain, intent and slot in SLU has not been studied properly, despite its usefulness for various downstream tasks such as contextual cross-domain ranking in natural language understanding~\cite{N18-3003}.


Learning distributed representations of words has been studied extensively in the past few years~\cite{mikolov2013distributed, pennington2014glove, Peters2018DeepCW}. The learned representations of words provide machines with better description  of the word context, resulting in notable improvements on a variety of downstream applications, such as domain classification~\cite{Kim2018EfficientLD}, information retrieval~\cite{ganguly2015word}, document classification~\cite{sebastiani2002machine}, question answering~\cite{tellex2003quantitative}, named entity recognition\cite{turian2010word}, and parsing~\cite{socher2013parsing}. Especially, one can leverage the word representations to achieve higher accuracy in domain classification, intent determination, and slot filling in SLU. Similarly, distributed representations for domain, intent, and slot can be helpful for downstream applications as well. For instance, in order to better support a large-scale domain classification, where tens of thousands of domains are available in SLU, a novel framework consisting of shortlisting candidate domains followed by contextual reranking has been recently proposed~\cite{Kim2018EfficientLD, N18-3003}. The Shortlister~\cite{Kim2018EfficientLD} first picks top $k$ candidate domains which are most likely to handle a given utterance, and then the contextual reranker~\cite{N18-3003} attempts to predict the best domain by utilizing various signals such as NLU scores for intent determination and named entity recognition, embedding vectors for the predicted domain/intent/slot, and contextual signals including time, location, customer's ratings, per domain usage history, and so on.

Domain, intent, and slot are structured hierarchically and there is inherent dependency between them. That is, a domain can be viewed as a group of intents which belong to the domain, and an intent consists of one or multiple slots that define semantic keywords to interpret the intent. The hierarchical information is able to bring us strong signals to properly map them to vectors in a low dimensional embedding space. Inspired by graph convolutional network~\cite{Kipf2016SemiSupervisedCW, hamilton2017inductive}, we propose a framework that learns coupled embeddings of the three SLU components by sampling and aggregating the embeddings of sub-class components. Specifically, slot representations are aggregated and then transformed into the intent vector space to define the representation of the intent to which the slots belong. The intent representations learned in that way are successively aggregated to define the representation of the domain in which the intents are specified.


We present our experimental results that empirically show how the learned representations are practically helpful for downstream applications and how well the proposed model is able to capture semantic closeness or distance between domains, intents, or slots. As one of the downstream applications for which the learned representations could be useful, we consider the contextual cross-domain reranking method for the problem of domain classification~\cite{N18-3003} and show that the classification accuracy can be affected by the quality of the representations of domain, intent, and slot.
\section{MODEL DESCRIPTION}
\label{sec:model_description}
In this section, we describe our proposed method to learn representations for SLU components - domain, intent, and slot, in detail.

\subsection{Problem Definition}
Suppose we are given a bunch of utterances, each of which is labeled with a domain $d_l, \forall l = 1, \dots, |D|$, an intent $i_{l,m}, \forall m = 1, \dots, |I_l|$, and consists of a sequence of words tagged with slots $s_{l,m,n}, \forall n = 1, \dots, |S_{l,m}|$, where $D$ is a set of all domains, $I_l$ is a set of intents which belong to the domain $d_l$, and $S_{l,m}$ is a set of slots used for the intent $i_{l,m}$. The goal is to learn representations of the three components in a low-dimensional continuous vector space,  $\mathbf{d} \in \mathds{R}^{k_d}, \mathbf{i} \in \mathds{R}^{k_i}$, and $\mathbf{s} \in \mathds{R}^{k_s}$, where $k_d$, $k_i$, and $k_s$ are the dimensionality of domain, intent, and slot embeddings, respectively. The learned representations are supposed to be able to capture common semantics shared by each component and distinguish specific features across different components. In other words, the embeddings that are semantically similar to each other should be located more closely to each other rather than others not sharing common semantics in the embedding space.

We do not use full subscripts for each component in the following sections if they are obvious. Section~\ref{sec:embedding_generation}, \ref{sec:aggregate_slots}, and \ref{sec:aggregate_intents} describe the embedding generation, or forward propagation algorithm, which assumes that the model has already been trained and that the parameters are fixed.

\subsection{Embedding Generation using Hierarchical Structure}
\label{sec:embedding_generation}
\begin{figure}[t]
	\centering
    \includegraphics[width=0.45\textwidth]{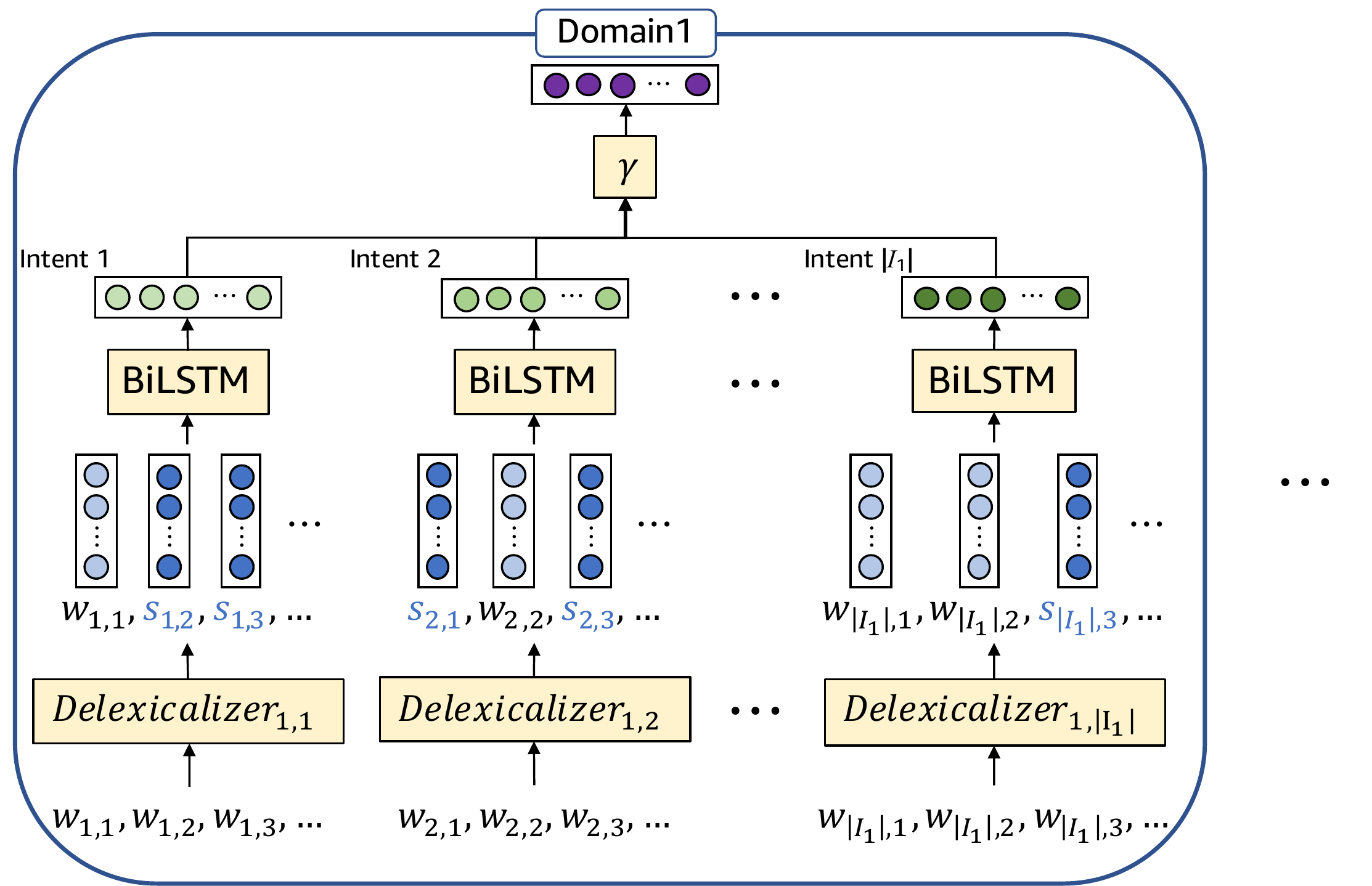}
    \caption{Model Architecture: This illustrates how the domain $d_1$, its intents, and the slots under the intents learn their embeddings using hierarchical structure along with aggregator functions. $\gamma$ is an element-wise max operator.}
    \label{fig:model_architecture}
\end{figure}
Each domain in $D$ is associated with a set of intents that define functionalities supported in the domain. For example, the domain \texttt{Music} may include the intents \textit{PlayMusicIntent}, \textit{SearchMusicIntent}, \textit{BuyMusicIntent}, and so on. Also, each intent consists of slots such as \textit{artist}, \textit{title}, and \textit{playlist}, and so on. That is, all the three components establish hierarchical structure from domains, through intents, to slots. The idea behind our proposed method to learn domain, intent, and slot embeddings lies in the hierarchy. The vector representation $\mathbf{d}$ of a domain can be obtained from aggregating the vector representations $\mathbf{i}$ of its intents, and likewise, the vector representation $\mathbf{i}$ of an intent can be obtained from aggregating the vector representations $\mathbf{s}$ of words and slots in an utterance. We explain further details in the following sections~\ref{sec:aggregate_slots} and~\ref{sec:aggregate_intents} 

\subsection{Aggregating Words and Slots}
\label{sec:aggregate_slots}
Given an utterance consisting of a sequence of words
\begin{align*}
	(w_1;s_1, w_2;s_2, \dots, w_t;s_t)
\end{align*}
where $s$ is a slot label for each word, we first de-lexicalize it by replacing words with a meaningful slot with their corresponding slots and leaving words with no associated slot, referred to as \textit{other}, as they are. For instance, if we are given an utterance \textit{``play;other Thriller;song of;other Michael;artist Jackson;artist''}, then it is de-lexicalized into \textit{``play song of artist''}. Pre-trained word embeddings are used to initialize learnable parameters for each word embedding. For slot embeddings, we identify in advance which words are used for each slot from the entire set of utterances and take the average of pre-trained embeddings of the words for an initial slot embedding. Let $\mathbf{x} = (\mathbf{x_1}, \mathbf{x_2}, \dots, \mathbf{x_t})$ denote such sequence of embeddings. Note that every intent comes with various patterns of utterances, which means that the embedding of an intent needs to represent all the different utterances. Thus, for every iteration in the model training, we draw sample utterances randomly, aggregate information about words and slots for each utterance, and then again take the average of the aggregated information from all the utterances.

Since it is important to learn sequential patterns from words and slots in an utterance, we adopt Bidirectional Long Short Term Memory networks (BiLSTM) as the function of aggregating information from words and slots. The aggregated information is propagated to an intent level to obtain intent embeddings through nonlinear transformation. That is,

\begin{equation}
	\mathbf{i} = \sigma (\mathbf{W_i} \times \text{AVG} (\{\text{BiLSTM} (\mathbf{x}), \forall \mathbf{x} \in N(i)\}))
\end{equation}
where $\sigma$ is a non-linear activation function, $\mathbf{W_i}$ is a matrix of weight parameters, AVG is an element-wise mean function over vectors, and $N(i)$ is a set of randomly drawn sample utterances which belong to the intent $i$. For each of all intents, we learn intent embeddings through the operation above.

\subsection{Aggregating Intents}
\label{sec:aggregate_intents}
\begin{figure*}
	\centering
    \begin{subfigure}[t]{0.3\textwidth}
    	\includegraphics[width=\textwidth]{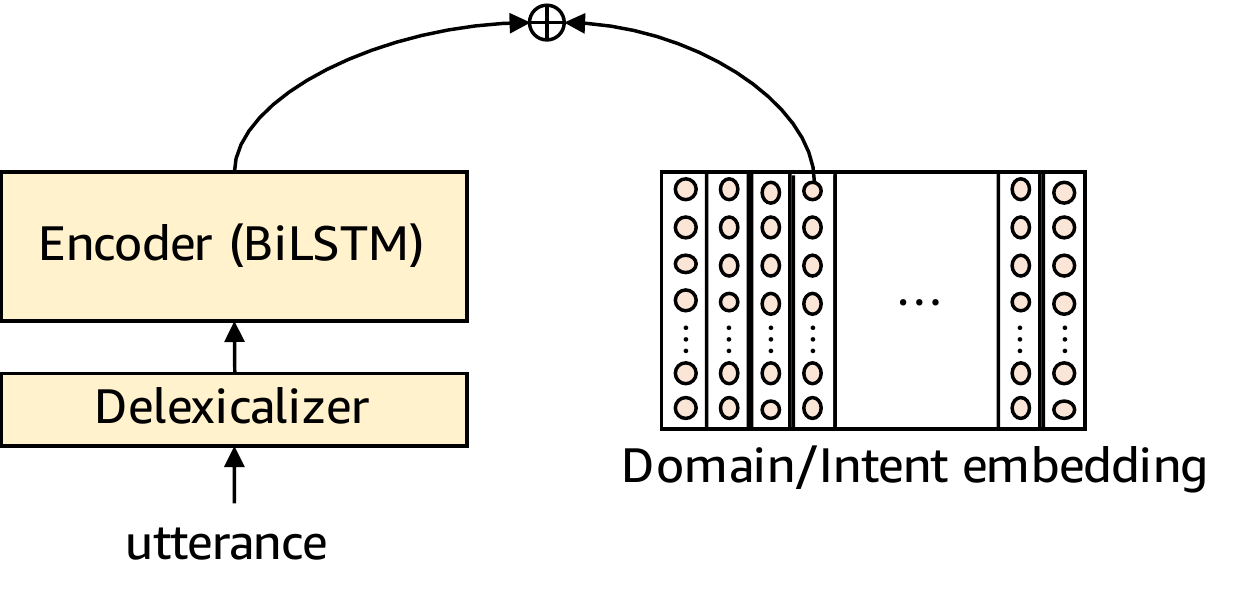}
        \caption{Baseline 1}
        \label{fig:baseline_1}
    \end{subfigure}
    ~
    \begin{subfigure}[t]{0.3\textwidth}
    	\includegraphics[width=\textwidth]{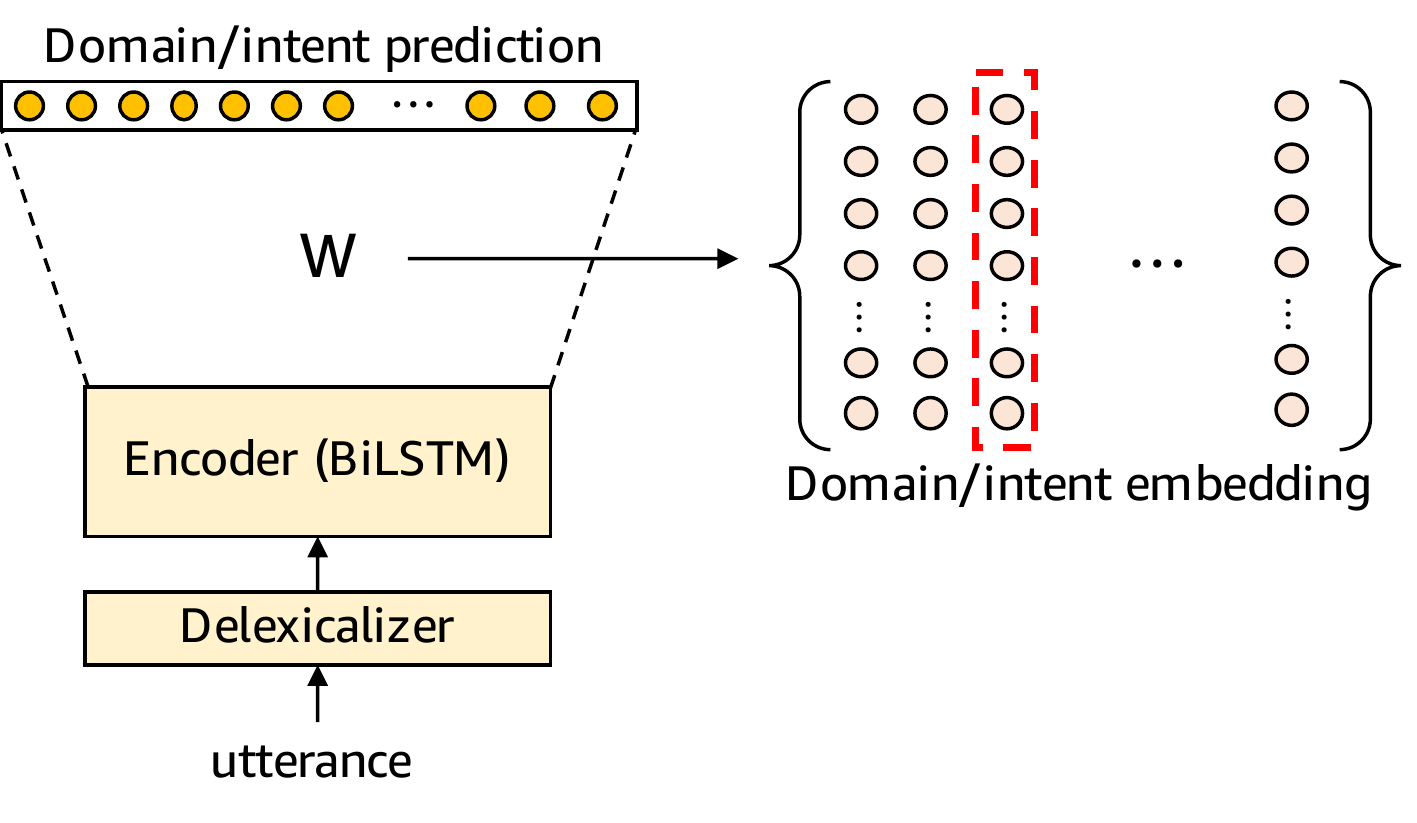}
        \caption{Baseline 2}
        \label{fig:baseline_2}
    \end{subfigure}
    ~
    \begin{subfigure}[t]{0.3\textwidth}
    	\includegraphics[width=\textwidth]{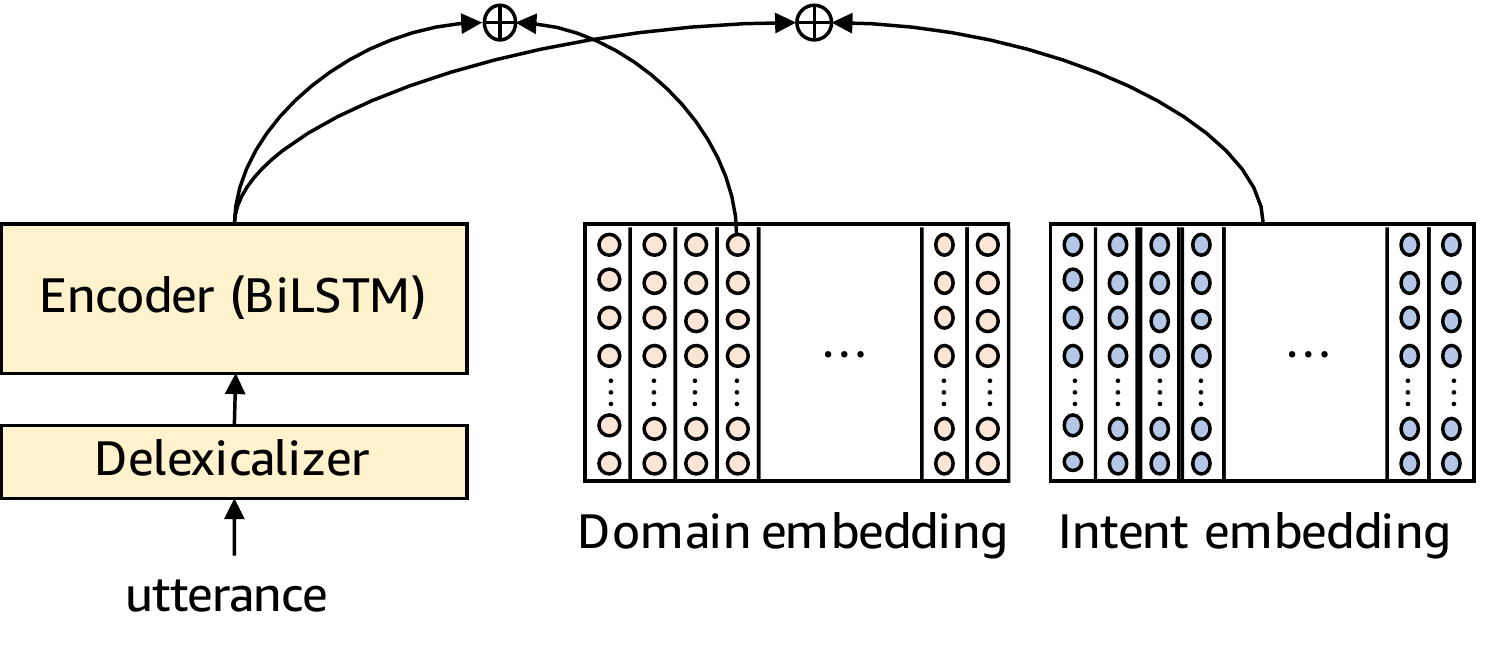}
        \caption{Baseline 3}
        \label{fig:baseline_3}
    \end{subfigure}
    \caption{Three competing baseline methods for learning domain/intent/slot embeddings.}
    \label{fig:baselines}
\end{figure*}

Once we obtain intent embeddings by aggregating information from words and slots, then those learned intent embeddings are aggregated and propagated into the domain level. As an intent can be considered as a semantic group of corresponding words and slots, a domain can be semantically defined as a group of intents that are used as a user's specific request to a domain or a domain's functionality in SLU.

Assume, for a particular domain $d$, we already have the embeddings of $d$'s intents, $I_d = {\mathbf{i_1}, \mathbf{i_2}, \dots, \mathbf{i_m}}$ learned from their words and slots. In order to aggregate the information from the intents of the domain $d$, we can use different aggregator functions such as average and max pooling. Taking an average acts as summarizing the information of all intents, and max pooling has the effect of taking only the most representative information from the intent embeddings. If an average is used for aggregation, then generating the domain embedding can be formally written as follows,
\begin{equation}
	\mathbf{d} = \sigma (\mathbf{W_d} \times \displaystyle \sum_{\mathbf{i} \in I_d} \frac{\mathbf{i}}{|I_d|})
\end{equation}
and if the max pooling is employed as the aggregator function, then the domain embedding $\mathbf{d}$ is obtained as follows,
\begin{equation}
	\mathbf{d} = \sigma (\mathbf{W_d} \times \gamma (\{(\mathbf{W_{pool}} \times \mathbf{i} + \mathbf{b}), \forall \mathbf{i} \in I_d\}))
\end{equation}
where $\gamma$ is an element-wise max operator.  In principle, the function applied before the max pooling can be an arbitrarily deep multi-layer perceptron, but we focus on simple single-layer architectures in this work. This approach is inspired by recent advancements in applying neural network architectures to learn over general point sets. Intuitively, the multi-layer perceptron can be thought of as a set of functions that compute features for each of the intent representations in the $d$'s intent set. By applying the max-pooling operator to each of the computed features, the model effectively captures different aspects of the intent set.

\subsection{Learning the Model Parameters}
Starting from pre-trained word embeddings and slot embeddings initialized from the word embeddings, we successively obtain intent embeddings, and domain embeddings using aggregated information in the hierarchical structure. Then, each domain defined in SLU has its own vector representation, which can be used for the task of domain prediction. This task-specific supervised learning is expected to optimize the set of learnable parameters in the model. Given $|D|$ different domains $(d_1, d_2, \dots, d_{|D|})$ and their corresponding embeddings $(\mathbf{d_1}, \mathbf{d_2}, \dots, \mathbf{d}_{|D|})$ in SLU, a multi-layer perceptron network is used for each domain to perform domain predictions. For a set of sampled utterances belonging to the domain $d_l$, the prediction scores over different domains are,
\begin{equation}
	\mathbf{z} = \text{softmax} (\mathbf{V_2} \times \sigma (\mathbf{V_1} \times \mathbf{d}_l + \mathbf{b_1}) + \mathbf{b_2})
\end{equation}
then the prediction loss for the domain $d_l$ is,
\begin{equation}
	\mathcal{L}_l = - \displaystyle \sum_{p=1}^{|D|} \mathds{1} (d_l = p) \log (\mathbf{z}_p)
\end{equation}
where $\mathds{1}$ is an indicator function that returns 1 if the domain label is $p$, otherwise 0. Since we assume there are $l$ different domains defined in SLU, we make $l$ separate domain predictions and put all their losses into the same objective function, as follows,
\begin{equation}
	\argmin_{\Theta} \mathcal{L}_1 + \mathcal{L}_2 + \cdots + \mathcal{L}_l
\end{equation}
where $\Theta$ is a set of all the learnable weight parameters we have discussed so far.

\section{EXPERIMENTS}
\label{sec:experiments}
\subsection{Datasets}
We demonstrate the effectiveness of the learned embeddings on a suite of Alexa domains~\cite{kumar2017just}, a large-scale real-world SLU system, through contextual cross-domain reranking task~\cite{N18-3003}. This dataset consists of 246,000 user utterances over 17 domains, 246 intents, and 3,409 slots.

\subsection{Baselines}
\label{sec:baselines}
To compare with our proposed method, we use three different baseline methods. Since there is no existing work that was proposed to perform exactly the same representation learning task, we have designed simple but effective methods, which are illustrated in Figure~\ref{fig:baselines}.

\noindent \textbf{Baseline 1} Holding one lookup parameters for word embeddings and the other lookup parameters for domain/intent embeddings, a sequence of words are first replaced with a sequence of words/slots using de-lexicalizer and then encoded into a vector representation by BiLSTM. Then, we take the dot product between the utterance embedding $\mathbf{u}$ and the corresponding domain embedding $\mathbf{d}$ or intent embedding $\mathbf{i}$ which is initialized with random values at first. We optimize the model using a binary log loss. Formally, for learning slot and domain embeddings,
\begin{equation}
	\mathcal{L}_d = - \log(\sigma(\mathbf{u}^{\top} \mathbf{d})) - Q \cdot \mathds{E}_{d_n \sim P_n(d)} \log(\mathbf{u}^{\top} \mathbf{d_n})
    \label{eq:baseline1_domain}
\end{equation}
and for learning slot and intent embeddings,
\begin{equation}
	\mathcal{L}_i = - \log(\sigma(\mathbf{u}^{\top} \mathbf{i})) - Q \cdot \mathds{E}_{i_n \sim P_n(i)} \log(\mathbf{u}^{\top} \mathbf{i_n})
    \label{eq:baseline1_intent}
\end{equation}
where $\sigma$ is the sigmoid function, $Q$ defines the number of negative samples, and $P_n$ is a negative sampling distribution. Once mode training is done, the lookup tables represent slot and domain/intent embeddings.

\noindent \textbf{Baseline 2} As done in Baseline 1, an input sequence of words is transformed into a sequence of words and slots and then is consumed by BiLSTM to produce its utterance embedding. We use the encoded vector directly to predict a domain/intent using a multi-layer perceptron network. Then, the weight matrix between the last hidden layer and the output layer contains the domain/intent representations as its column vectors. We extract those vectors after model training.

\noindent \textbf{Baseline 3} The domain/intent/slot embeddings all are jointly learned together in this baseline. After obtaining an utterance embedding $\mathbf{u}$ from BiLSTM, we take two dot products: one is between $\mathbf{u}$ and $\mathbf{d}$ and the other is between $\mathbf{u}$ and $\mathbf{i}$. The model's loss function is the sum of Equation~\ref{eq:baseline1_domain} and Equation~\ref{eq:baseline1_intent},
\begin{equation}
	\mathcal{L}_{d+i} = \mathcal{L}_d + \mathcal{L}_i
\end{equation}

Note that both Baseline 1 and 2 produce different sets of slot embeddings depending on which component is used in the objective function, between domain and intent, whereas Baseline 3 learns all the three embeddings simultaneously based on joint loss. We found out that no matter which component between domain and intent is used in Baseline 1 and Baseline 2, the resulting slot embeddings do not make significant differences in terms of the performance on the downstream task, contextual cross-domain reranking. We report the experimental results obtained when using slot embeddings learned along with domain embeddings in case of Baseline 1 and 2. 

\subsection{Evaluation Tasks}
We consider one downstream task to evaluate how well domain/intent/slot are represented in a low-dimensional vector space, using our proposed method. As mentioned earlier, the learned representations for domain/intent/slot can be greatly useful for a downstream application, the task of contextual cross-domain reranking~\cite{N18-3003}. In order to efficiently and successfully perform a large-scale domain classification where tens of thousands of domains are available, one may be required to first pick top candidates by utilizing only utterance text and then to rerank them based on richer contextual signals including estimated domain/intent/slot prediction scores, their embeddings, and other domain and/or user related meta data. Due to the sparsity in the domain/intent/slot embedding space, it is important to pre-train their embeddings in prior to performing the reranking task. We use the \textit{Shortlister} model~\cite{Kim2018EfficientLD} which is one of the state-of-the-art approaches to domain classification that has been recently proposed in order to first select the top candidates and then make final predictions using the \textit{HypRank} model~\cite{N18-3003} with randomly initialized or pre-trained domain, intent, and slot embeddings of the top candidates.


\subsection{Impact on Contextual Cross-Domain Reranking}
\begin{table}
	\centering
    \small
	\begin{tabular}{| c | c | c | c | c |}
    	\hline
    	Model & $d=50$ & $d=100$ & $d=200$ & $d=300$ \\ \hhline{|=|=|=|=|=|}
        SL & 0.832 & 0.832 & 0.832 & 0.832 \\ \hline
        SL+HR\_rand & 0.891 & 0.896 & 0.895 & 0.883 \\ \hline
        SL+HR\_baseline1 & 0.915 & 0.924 & 0.922 & 0.913 \\ \hline
        SL+HR\_baseline2 & 0.904 & 0.908 & 0.905 & 0.903 \\ \hline
        SL+HR\_baseline3 & 0.913 & 0.919 & 0.920 & 0.915 \\ \hline
        SL+HR\_hierarchy & \textbf{0.928} & \textbf{0.938} & \textbf{0.933} & \textbf{0.924} \\ \hline
    \end{tabular}
    \caption{Domain classification accuracy over different embedding dimensionalities. SL represents \textit{Shortlister} model without reranking, SL+HR\_rand is \textit{Shortlister} followed by \textit{HypRank} with randomly initialized embeddings, HR\_baseline1, HR\_baseline2, and HR\_baseline3 are the \textit{HypRank} models using the embedding obtained from the three different baselines, and HR\_hierarchy is the \textit{HypRank} using the embeddings learned by our proposed method based on max pooling.}
    \label{tab:reranking}
\end{table}

Table~\ref{tab:reranking} presents the domain classification accuracy of various approaches over different embedding dimensionalities. It is obvious that not only reranking with further contextual signals is helpful for improving the domain classification accuracy but also the domain/intent/slot embeddings can affect the model performance positively. If only \textit{Shortlister} is used for domain classification, then the model achieves 83.2\% accuracy. The contextual reranker, \textit{HypRank}, can improve the accuracy to different extents depending on how the domain/intent/slot embeddings used by \textit{HypRank} are initialized. If we just initialize the embeddings with random real values and optimize them through model training, the classification accuracy goes up to 0.896. On the other hand, pre-trained embeddings can make the model predictions more accurate rather than embeddings with random initial values. Especially, when the embeddings are learned by our proposed model, the contextual reranker outperforms others with embeddings learned by baselines that are discussed in Section~\ref{sec:baselines}. The embeddings learned by the proposed method can increase the accuracy up to 0.938, while the embeddings learned by the baseline methods make \textit{HypRank} achieve the accuracy in the range of 0.903 $\sim$ 0.924. It proves that the hierarchical structure among domains, intents, and slots is critical to extract inherent relationships between them and leads to creating well represented embeddings. Additionally, all the different approaches tend to optimize the embeddings best when the dimensionality of the embedding space is 100, and increasing the dimensionality can make a negative effect on the reranking performance.

\section{CONCLUSION}
\label{sec:conclusion}
In this paper, we propose a novel representation learning method for domain, intent, and slot in Spoken Language Understanding system. It exploits hierarchical information that resides in between the three components. Domain embedding is obtained from aggregating intent embeddings, and intent embedding is defined by the aggregation of slot embeddings along with words appearing in an utterance, successively. The aggregated information provides deeper contextual semantics for the upper-level components, which results in richer representations. The experimental results demonstrate the effectiveness of our proposed method in the downstream task, contextual cross-domain reranking. The proposed method is also expected to be able to easily apply to performing the SLU tasks such as domain classification, intent determination, and slot filling, directly.

\bibliographystyle{IEEEbib}
\bibliography{references}

\end{document}